\documentclass{svproc}
\usepackage{graphicx}%
\usepackage{multirow}%
\usepackage{amsmath,amssymb,amsfonts}%
\usepackage{mathrsfs}%
\usepackage[title]{appendix}%
\usepackage{xcolor}%
\usepackage{textcomp}%
\usepackage{manyfoot}%
\usepackage{booktabs}%
\usepackage{algorithm}%
\usepackage{algorithmicx}%
\usepackage{algpseudocode}%
\usepackage{listings}%
\usepackage{url}
\usepackage{float}
\usepackage{bm}
\usepackage{enumerate}
\usepackage{color, colortbl}
\usepackage{booktabs, multirow}
\definecolor{darkgreen}{rgb}{0, 0.5, 0}
\definecolor{red}{rgb}{1, 0, 0}
\definecolor{purple}{rgb}{0.5, 0, 0.5}
\usepackage{chngpage}
\usepackage{comment}
\usepackage{mfirstuc}
\usepackage{mathtools}
\usepackage{lettrine}
\usepackage{pxfonts}

\newcommand\iid{\textit{i.i.d.}}

\newcommand\ie{\textit{i.e.}}
\newcommand\eg{\textit{e.g.,}}

\newcommand\wrt{\textit{w.r.t.}}
\newcommand\etc{\textit{etc.}}

\newcommand{\norm}[1]{\left\lVert#1\right\rVert}
\newcommand{\Rbb}{{\mathbb{R}}}
\newcommand{\abs}[1]{\left| #1 \right|}
\newcommand{\beq}{\begin{equation}}
\newcommand{\eeq}{\end{equation}}
\newcommand{\beqnn}{\begin{equation*}}
\newcommand{\eeqnn}{\end{equation*}}
\newcommand{\beqy}{\begin{eqnarray}}
\newcommand{\eeqy}{\end{eqnarray}}
\newcommand{\beqynn}{\begin{eqnarray*}}
\newcommand{\eeqynn}{\end{eqnarray*}}
\newcommand{\bit}{\begin{itemize}}
\newcommand{\eit}{\end{itemize}}
\newcommand{\ben}{\begin{enumerate}}
\newcommand{\een}{\end{enumerate}}
\newcommand{\bex}{\begin{example}}
\newcommand{\eex}{\end{example}}

\newcommand{\diag}{\mathrm{diag}}

\newcommand{\trace}{\mathrm{trace}}

\renewcommand{\u}{\boldsymbol{u}}

\newcommand{\x}{{\boldsymbol{x}}}

\begin{document}
\mainmatter              
\title{When Do We Need Graph Neural Networks for Node Classification?}
\titlerunning{When Do We Need Graph Neural Networks for Node Classification?}  
%
\author{
Sitao Luan$^{1,2}$, Chenqing Hua$^{1,2}$, Qincheng Lu$^{1}$, Jiaqi Zhu$^{1}$, Xiao-Wen Chang$^{1}$, Doina Precup$^{1,2,3}$\\
\{sitao.luan@mail, chenqing.hua@mail, qincheng.lu@mail, jiaqi.zhu@mail, \\chang@cs, dprecup@cs\}.mcgill.ca\\
$^1$McGill University; $^2$Mila; $^3$DeepMind\\
}
\institute{}
\authorrunning{Sitao Luan et al.} 

\maketitle              

\begin{abstract}
Graph Neural Networks (GNNs) extend basic Neural Networks (NNs) by additionally making use of graph structure based on the relational inductive bias (edge bias), rather than treating the nodes as collections of independent and identically distributed (\iid) samples. Though GNNs are believed to outperform basic NNs in real-world tasks, it is found that in some cases, GNNs have little performance gain or even underperform graph-agnostic NNs. To identify these cases, based on graph signal processing and statistical hypothesis testing, we propose two measures which analyze the cases in which the edge bias in features and labels does not provide advantages. Based on the measures, a threshold value can be given to predict the potential performance advantages of graph-aware models over graph-agnostic models.
\end{abstract}

\section{Introduction}

\label{sec:introduction}
In the past decade, deep Neural Networks (NNs) \cite{lecun2015deep} have revolutionized many machine learning areas and one of their major strength is their capacity and effectiveness of learning latent representation from Euclidean data. Recently, the focus has been put on its applications on non-Euclidean data, \eg{} relational data or graphs. Combining with graph signal processing and convolutional neural networks \cite{lecun1998gradient}, numerous Graph Neural Networks (GNNs) have been proposed \cite{defferrard2016convolutional,hamilton2017inductive,velivckovic2018graph,kipf2016classification,luan2019break} that empirically outperform traditional neural networks on graph-based machine learning tasks, \eg{} node classification, graph classification, link prediction, graph generation, \etc

Nevertheless, growing evidence shows that GNNs do not always gain advantages over traditional NNs on relational data \cite{zhu2020generalizing,liu2020non,luan2022complete,lim2021large,luan2023addressing,luan2023graph}. In some cases, even a simple Multi-Layer Perceptron (MLP) can outperform GNNs by a large margin, \eg{} as shown in table \ref{tab:comparison_baselines_mlp}, MLP outperform baseline GNNs on \textit{Cornell, Wisconsin, Texas} and \textit{Film} and perform almost the same as baseline GNNs on \textit{PubMed, Coauthor CS} and \textit{Coauthor Phy}. This makes us wonder when it is appropriate to use GNNs. In this work, we explore an explanation and propose two proper measures to determine when to use GNNs for a node classification task.

A common way to leverage graph structure is to apply graph filters in each hidden layer of NNs to help feature extraction. Most existing graph filters can be viewed as operators that aggregate node information from its direct neighbors. Different graph filters yield different spectral or spatial GNNs. Among them, the most commonly used is the \textit{renormalized affinity matrix} \cite{kipf2016classification}, which corresponds to a low-pass (LP) filter \cite{maehara2019revisiting} mainly capturing the low-frequency components of the input, \ie the locally smooth features across the whole graph \cite{wu2019simplifying}.

The use of LP graph filters relies on the assumption that nodes tend to share attributes with their neighbors, a tendency called homophily \cite{mcpherson2001birds,hamilton2020graph} that is widely exploited in node classification tasks. GNNs that are built on the homophily assumption learn to assign similar labels to nodes that are closely connected \cite{zhou2004learning}, which corresponds to an assumption of intrinsic smoothness on latent label distribution. We call this kind of relational inductive bias \cite{battaglia2018relational} the edge bias. We believe it is a key factor leading to GNNs' superior performance over NNs' in many tasks. 

\begin{table}[htbp]
  \centering
  \small
  \caption{Accuracy (\%) Comparison of Baseline GNNs and MLP }
  \setlength{\tabcolsep}{3pt}
       \begin{tabular}{c|c|ccc|c|cc}
    \toprule
    \toprule
    \multirow{2}[2]{*}{Datasets\textbackslash{}Models} & MLP   & GCN   & GAT   & GraphSAGE & Baseline & Diff(MLP, & \multicolumn{1}{p{4.625em}}{Edge } \\
          & Acc   & Acc   & Acc   & Acc   &  Average & Baseline) & \multicolumn{1}{p{4.625em}}{Homophily} \\
    \midrule
    Cornell & 85.14 & 60.81 & 59.19 & 82.97 & 67.66 & \cellcolor[rgb]{ .439,  .678,  .278}\textbf{17.48} & 0.3 \\
    Wisconsin & 87.25 & 63.73 & 60.78 & 87.84 & 70.78 & \cellcolor[rgb]{ .439,  .678,  .278}\textbf{16.47} & 0.21 \\
    Texas & 84.59 & 61.62 & 59.73 & 82.43 & 67.93 & \cellcolor[rgb]{ .439,  .678,  .278}\textbf{16.66} & 0.11 \\
    Film  & 36.08 & 30.98 & 29.71 & 35.28 & 31.99 & \cellcolor[rgb]{ .439,  .678,  .278}\textbf{4.09} & 0.22 \\
    Chameleon & 46.21 & 61.34 & 61.95 & 47.32 & 56.87 & -10.66 & \textcolor[rgb]{ 1,  0,  0}{\textbf{0.23}} \\
    Squirrel & 29.39 & 41.86 & 43.88 & 30.16 & 38.63 & -9.24 & \textcolor[rgb]{ 1,  0,  0}{\textbf{0.22}} \\
    Cora  & 74.81 & 87.32 & 88.07 & 85.98 & 87.12 & -12.31 & 0.81 \\
    Citeseer & 73.45 & 76.70 & 76.42 & 77.07 & 76.73 & -3.28 & 0.74 \\
    Pubmed & 87.86 & 88.24 & 87.81 & 88.59 & 88.21 & \cellcolor[rgb]{ .502,  .502,  .502}\textbf{-0.35} & \textcolor[rgb]{ 1,  0,  0}{\textbf{0.80}} \\
    DBLP  & 77.39 & 85.87 & 85.89 & 81.19 & 84.32 & -6.93 & 0.81 \\
     Coauthor CS & 93.72 & 93.91 & 93.41 & 94.38 & 93.90 & \cellcolor[rgb]{ .502,  .502,  .502}\textbf{-0.18} & \textcolor[rgb]{ 1,  0,  0}{\textbf{0.81}} \\
     Coauthor Phy & 95.77 & 96.84 & 96.32 & OOM   & 96.58 & \cellcolor[rgb]{ .502,  .502,  .502}\textbf{-0.81} & \textcolor[rgb]{ 1,  0,  0}{\textbf{0.93}} \\
     AMZ Comp & 83.89 & 87.03 & 89.74 & 83.70 & 86.82 & -2.93 & 0.78 \\
    AMZ Photo  & 90.87 & 93.61 & 94.12 & 87.97 & 91.90 & -1.03 & \textcolor[rgb]{ 1,  0,  0}{\textbf{0.83}} \\
    \bottomrule
    \bottomrule
    \end{tabular}%
  \label{tab:comparison_baselines_mlp}%
\end{table}%

However, the existing homophily metrics are not appropriate to display the edge bias, \eg{} as shown in table \ref{tab:comparison_baselines_mlp}, MLP does not necessarily outperform baseline GNNs on some low homophily datasets (\textit{Chameleon and Squirrel}) and does not significantly underperform baseline GNNs on some high homophily datasets (\textit{PubMed,Coauthor CS,Coauthor Phy and AMZ Photo}). Thus, a metric that is able to indicate whether or not the graph-aware models can outperform graph-agnostic models is needed.

\paragraph{Contributions} In this paper, we discover that graph-agnostic NNs are able to outperform GNNs on a non-trivial set of graph datasets. To explain the performance inconsistency, we propose the Normalized Total Variation (NTV) and Normalized Smoothness Value (NSV) to measure the effect of edge bias on features and labels of an attribute graph. NSV leads us to conduct statistical hypothesis testings to examine how significant the effect of edge bias is. With the measures and analyses on $14$ real-world datasets, we are able to predict and explain the expected performance of graph-agnostic MLPs and GNN models.

The rest of this paper is organized as follows: In section \ref{sec:prelimiary_notation}, we introduce the notations and the background; In section \ref{sec:NTV_NSV}, we propose two measures of the effect of edge-bias and discuss their potential usage; In section \ref{sec:related_works}, we discuss the related works.

\section{Preliminaries}

\label{sec:prelimiary_notation}
After stating the motivations, in this section, we will introduce the used notations and formalize the idea. We use bold fonts for vectors (\eg{} $\bm{v}$). Suppose we have an undirected connected graph $\mathcal{G}=(\mathcal{V},\mathcal{E}, A)$ without bipartite component, where $\mathcal{V}$ is the node set with $\abs{\mathcal{V}}=N$; $\mathcal{E}$ is the edge set without self-loop; $A \in \mathbb{R}^{N\times N}$ is the symmetric adjacency matrix with $A_{ij}=1$ if and only if $e_{ij} \in \mathcal{E}$, otherwise $A_{ij}=0$; $D$ is the diagonal degree matrix, \ie{} $D_{ii} = \sum_j A_{ij}$ and $\mathcal{N}_i=\{j: e_{ij} \in \mathcal{E}\}$ is the neighborhood set of node $i$. A graph signal is a vector $\bm{x} \in \mathbb{R}^N$ defined on $\mathcal{V}$, where $x_i$ is defined on the node $i$. We also have a feature matrix ${X} \in \mathbb{R}^{N\times F}$ whose columns are graph signals and each node $i$ has a corresponding feature vector ${X_{i:}}$ with dimension $F$, which is the $i$-th row of ${X}$. We denote $Z\in \mathbb{R}^{N\times C}$ as label encoding matrix, where $Z_{i:}$ is the one hot encoding of the label of node $i$.

\subsection{Graph Laplacian and Affinity Matrix} \label{sec:laplacian_affinity_matrix}

The (combinatorial) graph Laplacian is defined as $L = D - A$, which is a Symmetric Positive Semi-Definite (SPSD) matrix \cite{chung1997spectral}. Its eigendecomposition gives $L=U\Lambda U^T$, where the columns of $U\in \Rbb^{N\times N}$ are orthonormal eigenvectors, namely the \textit{graph Fourier basis}, $\Lambda = \diag(\lambda_1, \ldots, \lambda_N)$ with $\lambda_1 \leq \cdots \leq \lambda_N$, and these eigenvalues are also called \textit{frequencies}. The graph Fourier transform of the graph signal $\x$ is defined as $\bm{x}_\mathcal{F} = U^{-1} \bm{x} = U^{T} \bm{x} = [\u_1^T\x, \ldots, \u_N^T\x]^T$, where $\bm{u}_i^T \bm{x}$ is the component of $\bm{x}$ in the direction of $\bm{u_i}$. 

Finding the eigenvalues and eigenvectors of a graph Laplacian is equivalent to solving a series of conditioned minimization problems relevant to function smoothness defined on $\mathcal{G}$. A smaller $\lambda_i$ indicates that basis $\bm{u_i}$ is a smoother function defined on $\mathcal{G}$ \cite{dakovic2019local}, which means any two elements of $\bm{u_i}$ corresponding to two connected nodes will be more similar. This property plays an important role in our paper.

Some graph Laplacian variants are commonly used, \eg{} the symmetric normalized Laplacian $L_{\text{sym}} = D^{-1/2} L D^{-1/2} = I-D^{-1/2} A D^{-1/2}$ and the random walk normalized Laplacian $L_{\text{rw}} = D^{-1} L = I - D^{-1} A$. $L_{\text{rw}}$ and $L_{\text{sym}}$ share the same eigenvalues that are in $[0,2)$, and their corresponding eigenvectors satisfy $\bm{u}_{\text{rw}}^i = D^{-1/2} \bm{u}_{\text{sym}}^i$. 

The affinity (transition) matrices can be derived from the Laplacians, \eg{} $A_\text{rw} = I - L_\text{rw} = D^{-1} A$, $A_\text{sym} = I-L_\text{sym} = D^{-1/2} A D^{-1/2}$ and  $\lambda_i(A_\text{rw}) = \lambda_i(A_\text{sym}) = 1- \lambda_i(A_\text{sym}) = 1- \lambda_i(A_\text{rw}) \in (-1,1]$. \cite{kipf2016classification} introduced the renormalized affinity and Laplacian matrices as $\hat{A}_\text{sym} = \tilde{D}^{-1/2} \tilde{A} \tilde{D}^{-1/2},\ \hat{L}_{\text{sym}} = I - \hat{A}_\text{sym}$ , where $\tilde{A} \equiv A+I, \tilde{D} \equiv D+I$. It essentially adds a self-loop and is widely used in Graph Convolutional Network (GCN) as follows,
\begin{equation}
    \label{eq:gcn_original}
   Y = \text{softmax} (\hat{A}_\text{sym} \; \text{ReLU} (\hat{A}_\text{sym} {X} W_0 ) \; W_1 )
\end{equation}
where $W_0 \in \Rbb^{F\times F_1}$ and $W_1 \in \Rbb^{F_1\times O}$ are parameter matrices. GCN can learn by minimizing the following cross entropy loss
\begin{equation}
\label{eq:cross_entropy_loss}
     \mathcal{L}  = -\trace(Z^T \log Y).
\end{equation}
The random walk renormalized matrices $\hat{A}_{\text{rw}} = \tilde{D}^{-1} \tilde{A}$ can also be applied to GCN and $\hat{A}_{\text{rw}}$ shares the same eigenvalues as $\hat{A}_{\text{sym}}$. The corresponding Laplacian is defined as $\hat{L}_{\text{rw}} = I - \hat{A}_\text{rw}$ Specifically, the nature of random walk matrix makes $\hat{A}_{\text{rw}}$ behaves as a mean aggregator $(\hat{A}_{\text{rw}} \bm{x})_i = \sum_{j\in\{\mathcal{N}_i \cup i\}} {x}_j/(D_{ii}+1)$ which is applied in \cite{hamilton2017inductive} and is important to bridge the gap between spatial- and spectral-based graph convolution methods.

\section{Measuring the Effect of Edge Bias}

\label{sec:NTV_NSV}
In this section, we will derive two measures for the effect of edge bias and conduct hypothesis testing for the effect. We analyze the behaviors of these measures and apply them on $14$ real world datasets. The measurement results are used to predict the potential performance differences between GNNs and MLPs.

\subsection{Normalized Total Variation (NTV) \& Normalized Smoothness Value (NSV) for Measuring Edge Bias}

\paragraph{NTV} Graph Total Variation (GTV) is a quantity to characterize how much graph signal varies \wrt{} graph filters and is defined  as follows \cite{chen2015signal,ahmed2017graph},
\begin{align*}
    GTV(\bm{x}) = \norm{\bm{x} - \hat{A} \bm{x}}_p^p
\end{align*}
where $\hat{A}$ generally represents normalized or renormalized filters, the $\ell_p$-norm can be replaced by the Frobenius norm when we measure a matrix $X$. $GTV$ generally measures the utility of the edge bias by gauging the distance between node features and its aggregated neighborhood features. To eliminate the influence of the magnitude of $\bm{x}$ or $X$ and make it comparable, we define Normalized Total Variation (NTV) as follows,
\begin{equation}
    \text{NTV}(\bm{x}) = \frac{\norm{\bm{x} - \hat{A} \bm{x}}_2^2}{2 \norm{\bm{x}}_2^2}, \ \text{NTV}(X) = \frac{\norm{X - \hat{A} X}_F^2}{2 \norm{X}_F^2}
\end{equation}
the division of factor 2 guarantees that $0\leq \text{NTV} \leq 1$. A small NTV value implies $\bm{x} \approx \hat{A} \bm{x}$ or $X \approx \hat{A} X$.

\paragraph{NSV} 
Even when the features of the node resemble its aggregated neighborhood, it does not necessarily mean that the average pairwise attribute distance of connected nodes is smaller than that of unconnected nodes. Based on this argument, we define Normalized Smoothness Value (NSV) as a measure of the effect of the edge bias.

The total pairwise attribute distance of connected nodes is equivalent to the Dirichlet energy of $X$ on $\mathcal{G}$ as follows,
\begin{align*}
    E_D^\mathcal{G}({X}) &= \sum\limits_{i \leftrightarrow j} \norm{X_{i:} - X_{j:}}_2^2 = \sum\limits_{i \leftrightarrow j} (\bm{e_i}-\bm{e_j})^T {X} {X}^T (\bm{e_i}-\bm{e_j})  = tr\left(\sum\limits_{i \leftrightarrow j} (\bm{e_i}-\bm{e_j})^T {X} {X}^T (\bm{e_i}-\bm{e_j}) \right) \\
    & = tr\left(\sum\limits_{i \leftrightarrow j} (\bm{e_i}-\bm{e_j}) (\bm{e_i}-\bm{e_j})^T {X} {X}^T \right) = \trace({X}^T L {X}).
\end{align*}
The total pairwise distance of unconnected nodes can be derived from the Laplacian $L^{C}$ of the complementary graph $\mathcal{G}^C$. To get $L^C$, we introduce the adjacency matrix of $\mathcal{G}^C$  as $A^{C} = (\bm{1}\bm{1}^T-I) -A$, its degree matrix $D^{C} = (N-1)I-D$, and $L^{C} = D^{C} - A^{C} = NI - \bm{1}\bm{1}^T - L$. Then, the total pairwise attribute distance of unconnected nodes (Dirichlet energy of $X$ on $\mathcal{G^C}$) is 
\begin{align*}
E_D^{\mathcal{G}^C}({X})\!=\!\trace\left({X}^T L^C {X}\right)\!=\!\trace\left({X}^T (NI\!-\!\bm{1}\bm{1}^T ) {X}\right) - E_D^\mathcal{G}({X})
\end{align*}
$E_D^\mathcal{G}({X})$ and $E_D^{\mathcal{G}^C}({X})$, are non-negative and are closely related to sample covariance matrix (see appendix \ref{appendix:nsv_and_covariance_matrix} for details) as follows,
\begin{align*}
E_D^\mathcal{G}({X}) + E_D^{\mathcal{G}^C}({X}) &= \trace\left({X}^T (NI - \bm{1}\bm{1}^T ) {X}\right) = N(N-1) \cdot \trace \left(\text{Cov}(X)\right).
\end{align*}

Since $\trace \left(\text{Cov}(X)\right)$ is the total variation in $X$, we can say that the total sample variation can be decomposed in a certain way onto $\mathcal{G}$ and $\mathcal{G}^C$ as $E_D^\mathcal{G}({X})$ and $E_D^\mathcal{G^C}({X})$.
Then, the average pairwise distance (variation) of connected nodes and unconnected nodes can be calculated by normalizing $E_D^\mathcal{G}({X})$ and $E_D^\mathcal{G^C}({X})$,
\begin{equation}  
 E_{\text{N}}^\mathcal{G}({X}) = \frac{E_D^\mathcal{G}({X}) }{2\abs{\mathcal{E}}}, \ \ E_{\text{N}}^\mathcal{G^C}({X}) = \frac{E_D^\mathcal{G^C}({X})}{N(N-1)-2\abs{\mathcal{E}}}
\end{equation}
and the Normalized Smoothness Value (NSV) is defined as
\begin{equation}
    \begin{aligned}
    &\text{NSV}^\mathcal{G}(X) = \frac{E_{\text{N}}^\mathcal{G}({X})}{E_{\text{N}}^\mathcal{G}({X}) + E_{\text{N}}^\mathcal{G^C}({X})}. 
\end{aligned}  
\end{equation}

We can see that $0 \leq \text{NSV}^\mathcal{G}(X) \leq 1$ and it can be used to interpret the edge bias: (1) For labels $Z$, NSV$^\mathcal{G}(Z) \geq 0.5$ means that the proportion of connected nodes that share different labels is larger than that of unconnected nodes, which implies that edge bias is harmful for $Z$ and the homophily assumption is invalid; (2) For features $X$, NSV$^\mathcal{G}(X) \geq 0.5$ means that the average pairwise feature distance of connected nodes is greater than that of unconnected nodes, which suggests that the feature is non-smooth. On the contrary, small NSV$(Z)$ and NSV$(X)$ indicates that the homophily assumption holds and the edge bias is potentially beneficial.

The above analysis raises another question: how much does NSV deviating from 0.5 or what is the exact NSV to indicate the edge bias is statistically beneficial or harmful. In the following section, we study the problem from statistical hypothesis testing perspective and provide thresholds by the p-values.

\subsection{Hypothesis Testing for Edge Bias}

Consider the following distributions of labels and features,

For labels $Z$: 
\begin{itemize}
    \item $P_1= \mathbb{P}\left(Z_{i:} \neq Z_{j:} \big| e_{ij} \in \mathcal{E} \right)=$ The proportion of connected nodes that share different labels;
    \item $P_2=\mathbb{P} \left(Z_{i:} \neq Z_{j:} \big| e_{ij} \not\in \mathcal{E} \right)=$ The proportion of unconnected nodes that share different labels.
\end{itemize}
For features $X$: 
\begin{itemize}
    \item $D_1=\norm{X_{i:}-X_{j:}}_2^2 \; \big| e_{ij} \in \mathcal{E} =$ Distribution of pairwise feature distance of connected nodes;
    \item $D_2=\norm{X_{i:}-X_{j:}}_2^2  \; \big| e_{ij} \not\in \mathcal{E} =$ Distribution of pairwise feature distance of unconnected nodes.
\end{itemize}
Suppose $P_1, P_2, D_1, D_2$ follow:
\begin{align*}
    &P_1 \sim \text{Binom}(n_1,p_1),\ P_2 \sim \text{Binom}(n_2,p_2); \; D_1 \sim N(d_1,\sigma_1^2),\ D_2 \sim N(d_2,\sigma_2^2).
\end{align*}
Consider the hypotheses for labels
\begin{align*}
    H_0^{L}: p_1=p_2; \ H_1^L: p_1\neq p_2;\ H_2^L: p_1 \geq p_2; \ H_3^L: p_1 \leq p_2
\end{align*}
and hypotheses for features
\begin{align*}
    H_0^{F}: d_1=d_2; \ H_1^F: d_1\neq d_2;\ H_2^F: d_1 \geq d_2; \ H_3^F: d_1 \leq d_2
\end{align*}

To conduct the hypothesis tests, we use Welch's t-test for features and $\chi^2$ test for labels. We can see $E_{\text{N}}^\mathcal{G}({Z})$ and $E_{\text{N}}^{\mathcal{G}^C}({Z})$ are sample estimation of the mean $p_1$ and $p_2$ for label $Z$; $E_{\text{N}}^\mathcal{G}({X})$ and $E_{\text{N}}^\mathcal{G^C}({X})$ are sample estimation of mean $d_1$ and $d_2$ for $X$. Thus, the p-values of hypothesis tests can suggest if NSV statistically deviates from 0.5. The smoothness of labels and features can be indicated as follows,
\paragraph{For feature $X$:}
\begin{itemize}
    \item p-value($H_0^F$ vs $H_1^F$): $> 0.05$, $H_0^F$ holds, feature is non-smooth; $\leq 0.05$, to be determined.
     \item p-value($H_0^F$ vs $H_2^F$):  $\leq 0.05$, feature is statistically significantly non-smooth.
     \item p-value($H_0^F$ vs $H_3^F$):  $\leq 0.05$, feature is statistically significantly smooth.
\end{itemize}

\paragraph{For label $Z$:}
\begin{itemize}
    \item p-value($H_0^L$ vs $H_1^L$): $> 0.05$, $H_0^L$ holds, label is non-smooth; $\leq 0.05$, to be determined.
     \item p-value($H_0^L$ vs $H_2^L$): $\leq 0.05$, label is statistically significantly non-smooth.
     \item p-value($H_0^L$ vs $H_3^L$): $\leq 0.05$, label is statistically significantly smooth.
\end{itemize}
Results of hypothesis testing are summarized in Table \ref{tab:stats}. We can see that for the datasets where baseline GNNs underperform MLP, \textit{Cornell, Texas} and \textit{Wisconsin} has statistically significantly non-smooth labels and \textit{Film} has non-smooth labels. In these datasets, the edge bias will provide harmful information no matter the features are smooth or not. For other datasets, they have statistically significantly smooth labels, which means the edge bias can statistically provide benefits to the baseline GNNs and lead them to have superiority performance over MLP.

\begin{table*}[htbp]
  \centering
  \caption{Statistics of Datasets and the Performance Differences}
  \resizebox{\textwidth}{!}{
    \begin{tabular}{c|cc|ccc|cc|ccc|c}
    \toprule
    \toprule
    \multirow{2}[2]{*}{Datasets\textbackslash{}Measures} & \multicolumn{5}{c|}{Features}         & \multicolumn{5}{c|}{Labels}           & \multicolumn{1}{p{4.085em}}{Baseline Average -} \\
          & \multicolumn{1}{p{2.915em}}{  NTV} & \multicolumn{1}{p{2.915em}}{NSV} & \multicolumn{1}{p{4.085em}}{$H_0^F$ vs $H_1^F$} & \multicolumn{1}{p{4.085em}}{$H_0^F$ vs $H_2^F$} & \multicolumn{1}{p{4.085em}|}{$H_0^F$ vs $H_3^F$} & \multicolumn{1}{p{2.915em}}{  NTV} & \multicolumn{1}{p{2.915em}}{NSV} & \multicolumn{1}{p{4.085em}}{$H_0^L$ vs $H_1^L$} & \multicolumn{1}{p{4.085em}}{$H_0^L$ vs $H_2^L$} & \multicolumn{1}{p{4.085em}|}{$H_0^L$ vs $H_3^L$} & \multicolumn{1}{p{4.085em}}{MLP} \\
    \midrule
    Cornell & 0.33  & 0.48  & 0.00  & 1.00  & \cellcolor[rgb]{ .647,  .647,  .647}\textbf{0.00} & 0.33  & 0.53  & 0.0003	 & \cellcolor[rgb]{ .647,  .647,  .647}\textbf{0.00} & 1.00  & \textcolor[rgb]{ 1,  0,  0}{-17.48} \\
    Texas & 0.33  & 0.48  & 0.00  & 1.00  & \cellcolor[rgb]{ .647,  .647,  .647}\textbf{0.00} & 0.42  & 0.60  & 0.00  & \cellcolor[rgb]{ .647,  .647,  .647}\textbf{0.00} & 1.00  & \textcolor[rgb]{ 1,  0,  0}{-16.66} \\
    Wisconsin & 0.38  & 0.51  & \cellcolor[rgb]{ .647,  .647,  .647}\textbf{0.72} & 0.36  & 0.64  & 0.40  & 0.55  & 0.00  & \cellcolor[rgb]{ .647,  .647,  .647}\textbf{0.00} & 1.00  & \textcolor[rgb]{ 1,  0,  0}{-16.47} \\
    Film  & 0.39  & 0.50  & \cellcolor[rgb]{ .647,  .647,  .647}\textbf{0.19} & 0.90  & 0.10  & 0.37  & 0.50  & \cellcolor[rgb]{ .647,  .647,  .647}\textbf{0.05} & 0.97  & 0.03  & \textcolor[rgb]{ 1,  0,  0}{-4.09} \\
    \midrule
     Coauthor CS & 0.36  & 0.36  & 0.00  & 1.00  & \cellcolor[rgb]{ .647,  .647,  .647}\textbf{0.00} & 0.19  & 0.18  & 0.00  & 1.00  & \cellcolor[rgb]{ .647,  .647,  .647}\textbf{0.00} & \textcolor[rgb]{ 0,  .69,  .314}{0.18} \\
    Pubmed & 0.33  & 0.44  & 0.00  & 1.00  & \cellcolor[rgb]{ .647,  .647,  .647}\textbf{0.00} & 0.25  & 0.24  & 0.00  & 1.00  & \cellcolor[rgb]{ .647,  .647,  .647}\textbf{0.00} & \textcolor[rgb]{ 0,  .69,  .314}{0.35} \\
     Coauthor Phy & 0.35  & 0.36  & 0.00  & 1.00  & \cellcolor[rgb]{ .647,  .647,  .647}\textbf{0.00} & 0.16  & 0.09  & 0.00  & 1.00  & \cellcolor[rgb]{ .647,  .647,  .647}\textbf{0.00} & \textcolor[rgb]{ 0,  .69,  .314}{0.81} \\
     AMZ Photo  & 0.41  & 0.39  & 0.00  & 1.00  & \cellcolor[rgb]{ .647,  .647,  .647}\textbf{0.00} & 0.23  & 0.17  & 0.00  & 1.00  & \cellcolor[rgb]{ .647,  .647,  .647}\textbf{0.00} & \textcolor[rgb]{ 0,  .69,  .314}{1.03} \\
    \midrule
    AMZ Comp & 0.41  & 0.38  & 0.00  & 1.00  & \cellcolor[rgb]{ .647,  .647,  .647}\textbf{0.00} & 0.25  & 0.22  & 0.00  & 1.00  & \cellcolor[rgb]{ .647,  .647,  .647}\textbf{0.00} & \textcolor[rgb]{ 0,  .69,  .314}{2.93} \\
    Citeseer & 0.35  & 0.45  & 0.00  & 1.00  & \cellcolor[rgb]{ .647,  .647,  .647}\textbf{0.00} & 0.22  & 0.24  & 0.00  & 1.00  & \cellcolor[rgb]{ .647,  .647,  .647}\textbf{0.00} & \textcolor[rgb]{ 0,  .69,  .314}{3.28} \\
    DBLP  & 0.37  & 0.46  & 0.00  & 1.00  & \cellcolor[rgb]{ .647,  .647,  .647}\textbf{0.00} & 0.21  & 0.20  & 0.00  & 1.00  & \cellcolor[rgb]{ .647,  .647,  .647}\textbf{0.00} & \textcolor[rgb]{ 0,  .69,  .314}{6.93} \\
    Squirrel & 0.47  & 0.54  & 0.00  & \cellcolor[rgb]{ .647,  .647,  .647}\textbf{0.00} & 1.00  & 0.44  & 0.49  & 0.00  & 1.00  & \cellcolor[rgb]{ .647,  .647,  .647}\textbf{0.00} & \textcolor[rgb]{ 0,  .69,  .314}{9.24} \\
    Chameleon & 0.45  & 0.45  & 0.00  & 1.00  & \cellcolor[rgb]{ .647,  .647,  .647}\textbf{0.00} & 0.45  & 0.49  & 0.00  & 1.00  & \cellcolor[rgb]{ .647,  .647,  .647}\textbf{0.00} & \textcolor[rgb]{ 0,  .69,  .314}{10.66} \\
    Cora  & 0.38  & 0.47  & 0.00  & 1.00  & \cellcolor[rgb]{ .647,  .647,  .647}\textbf{0.00} & 0.20  & 0.19  & 0.00  & 1.00  & \cellcolor[rgb]{ .647,  .647,  .647}\textbf{0.00} & \textcolor[rgb]{ 0,  .69,  .314}{12.31} \\
    \bottomrule
    \bottomrule
    \end{tabular}%
    }
  \label{tab:stats}%
\end{table*}%
\subsection{Why NTV and NSV Work}

We explain why and how NTV and NSV can be used to explain the performance gain and loss of GNNs over graph-agnostic NNs. We simplify the explanation by removing the non-linearity as \cite{wu2019simplifying}. Let $\hat{A}$ denote a general filter with $\norm{\hat{A}}_2 = 1$ in GNNs.

\paragraph{NTV} When the NTV of node features $X$ and labels $Z$ are are small, it implies
\begin{equation} \label{eq:ntv_small_implication}
    \hat{A} X \approx X, \; \hat{A} Z \approx Z.
\end{equation}
The loss function of GNNs and MLP can be written as follows,
\begin{equation}
    \text{GNNs: } \min_W \norm{\hat{A}XW-Z}_F ,\; \text{MLP: } \; \min_W \norm{XW-Z}_F,
\end{equation}
where $W$ is the learnable parameter matrix. When $\hat{A} Z \approx Z$,
\begin{equation}
\begin{aligned}
 \min_W \norm{\hat{A}XW-Z}_F \approx \min_W \norm{\hat{A}XW-\hat{A}Z}_F \leq \min_W \norm{\hat{A}}_2 \norm{XW-Z}_F =\min_W  \norm{XW-Z}_F.
\end{aligned}
\end{equation}
This suggests that GNNs work more effectively than graph-agnostic methods when NTV$^\mathcal{G}(Z)$ is small. However, when labels are non-smooth on $\mathcal{G}$, a projection onto the column space of $\hat{A}$  will hurt the expressive power of the model. In a nutshell, GNNs potentially have stronger expressive power than NNs when NTV$^\mathcal{G}(Z)$ is small.

\paragraph{NSV} We first rewrite the softmax function as follows,
\begin{equation}
    \begin{aligned}
    Y & = \text{softmax} (\hat{A}  X W ) =\left(\exp({Y'}) \bm{1} \bm{1}^T \right)^{-1}  \odot \exp({Y'}) \\
    \end{aligned}
\end{equation}
where $Y' = \hat{A}  X W,\ \bm{1} \in \mathbb{R}^{C\times 1}$ and $C$ is the output dimension. The loss function \eqref{eq:cross_entropy_loss} can be written as
\begin{equation}
    \begin{aligned}
    \label{eq:nll_loss_explanation}
    \mathcal{L} & = -\trace\left(Z^T \hat{A} X W  \right) + \trace\left(\bm{1}^T  \log\left(\exp({Y'}) \bm{1} \right)  \right).
    \end{aligned}
\end{equation}
We denote $\tilde{X}=XW$ and consider $-\trace\left(Z^T \hat{A} X W  \right)$, which plays the main role in the above optimization problem. 
\begin{equation}
    \begin{aligned}
    \label{eq:adjacency_and_labels_relation}
    -\trace\left(Z^T \hat{A} X W  \right) = -\trace\left(Z^T \hat{A} \tilde{X} \right) = - \sum\limits_{i\leftrightarrow j}  \hat{A}_{ij} Z_{i:}\tilde{X}_{j:}^T.
    \end{aligned}
\end{equation}
To minimize $\mathcal{L}$, if $\hat{A}_{ij}\neq 0$, then $\tilde{X}_{j:}$ will learn to get closer to $Z_{i:}$ and this means: (1) If $Z_{i:}=Z_{j:}$, $\tilde{X}_{j:}$ will learn to approach to the unseen ground truth label $Z_{j:}$ which is beneficial; (2) If $Z_{i:}\neq Z_{j:}$, $\tilde{X}_{j:}$ tends to learn a wrong label, in which case the edge bias becomes harmful. Conventional NNs can be treated as a special case with only $\hat{A}_{ii} =1$, otherwise 0. So the edge bias has no effect on conventional NNs.

To evaluate the effectiveness of edge bias, NSV makes a comparison to see if the current edges in $\mathcal{E^G}$ have significantly less probability of indicating different pairwise labels than the rest edges. If NSV together with the p-value suggests that the edge bias is statistically beneficial, we are able to say that GNNs will obtain performance gain from edge bias; otherwise, the edge bias will have a negative effect on GNNs. NTV, NSV, p-values and the performance comparison of baseline models on 14 real-world datasets shown in Table \ref{tab:stats} are consistent with our analysis.

\section{Related Works}

\label{sec:related_works}
\paragraph{Smoothness (Homophily)} The idea of node homophily and its measures are mentioned in \cite{pei2020geom}
and defined as follows,
\begin{equation*}
 H_\text{node}(\mathcal{G}) = \frac{1}{|\mathcal{V}|} \sum_{v \in \mathcal{V}} 
    \frac{\big|\{u \mid u \in \mathcal{N}_v, Z_{u,:}=Z_{v,:}\}\big|}{d_v}
\end{equation*}
Or in \cite{zhu2020generalizing}, the edge homophily is defined as follows, 
\begin{equation*}
 H_\text{edge}(\mathcal{G}) = \frac{\big|\{e_{uv} \mid e_{uv}\in \mathcal{E}, Z_{u,:}=Z_{v,:}\}\big|}{|\mathcal{E}|}
\end{equation*}
To avoid sensitivity to imbalanced classes, the class homophily\cite{lim2021new} is defined as 
\begin{equation*}
    H_\text{class}(\mathcal{G}) \!=\! \frac{1}{C\!-\!1} \sum_{k=1}^{C}\bigg[h_{k}
    \!-\! \frac{\big|\{v \!\mid\! Z_{v,k} \!=\! 1 \}\big|}{N}\bigg]_{+}, \ \
h_{k}\! =\! \frac{\sum_{v \in \mathcal{V}} \big|\{u \!\mid\! Z_{v,k}\! =\! 1, u \in \mathcal{N}_v, Z_{u,:}\!=\!Z_{v,:}\}\big| }{\sum_{v \in \{v|Z_{v,k}=1\}} d_{v}}
\end{equation*}
where $[a]_{+}=\max (a, 0)$; $h_{k}$ is the class-wise homophily metric. The above measures only consider the label consistency of connected nodes but ignore the unconnected nodes. Stronger label consistency can potentially happen in unconnected nodes, in which case the edge bias is not necessarily beneficial for GNNs. Aggregation homophily \cite{luan2021heterophily,luan2022revisiting} tries to capture the post-aggregation node similarity and is proved to be better than the above homophily measures. But, it is not able to give a clear threshold value to determine when GNNs can outperform graph-agnostic NNs.

\paragraph{Connections and Differences among Terminologies}
We draw the connections and differences among edge bias, homophily/heterophily and smoothness/non-smoothness, which are frequently used in the literature that might cause confusion. Edge bias or homophily/smoothness assumption is a major and strong condition that is taken for granted when designing GNN models. When the homophily/smooth assumption holds, edge bias will have positive effects for training GNNs; On the contrary, when heterophily/non-smoothness assumption holds, edge bias will cause negative effects. The fact that, the current measures of homophily/heterophily do not consider unconnected nodes, poses chanllenges to fully examine the effect of edge bias or if homophily/heterophily assumption holds. The edge bias might cause some other problems, \eg{} over-smoothing \cite{li2018deeper}, loss of rank \cite{luan2019break} and training difficulty \cite{cong2021provable,luan2020training}, but we mainly discuss homophily/heterophily problem in this paper.
\section{Conclusion}
In this paper, we developed two measures, Normalized Total Variation (NTV) and Normalized Smoothness Value (NSV), which can predict and explain the expected performance of graph-agnostic MLPs and GNN models on graphs. These measures analyze the impact of edge bias on the features and labels of an attribute graph, helping to determine when graph-aware models will outperform graph-agnostic models. By conducting statistical hypothesis testing based on these measures, we are able to determine the threshold value for predicting the potential performance advantages of GNNs over NNs. Overall, our work contributes to a better understanding of the situations in which GNNs should be used, providing insights into the performance of GNNs compared to NNs on various real-world benchmark graph datasets.

\bibliography{references}
\bibliographystyle{abbrv}

\clearpage
\appendix

\section{Details of NSV and Sample Covariance Matrix}
\label{appendix:nsv_and_covariance_matrix}

The sample covariance matrix $S$ is computed as follows 
\begin{equation}
\begin{aligned}
{X} & =\left[\begin{array}{c}\bm{x}_{1:} \\ \vdots \\ \bm{x}_{N:} \end{array}\right], \ \ \bar{\bm{x}} = \frac{1}{N} \sum_{i=1}^{N} \bm{x}_{i:} = \frac{1}{N} \bm{1}^T {X}, \\ 
S & = \frac{1}{N-1} \left({X}-\bm{1} \bar{\bm{x}} \right)^{\top} \left({X}-\bm{1} \bar{\bm{x}} \right)
\end{aligned}
\end{equation}
It is easy to verify that
\begin{equation}
\begin{aligned}
S &= \frac{1}{N-1} \left({X}-\frac{1}{N} \bm{1}\bm{1}^T {X} \right)^{\top} \left({X}- \frac{1}{N} \bm{1} \bm{1}^T {X} \right) \\
& = \frac{1}{N-1} \left({X}^T {X} - \frac{1}{N} {X}^T \bm{1} \bm{1}^T {X} \right)\\
& = \frac{1}{N(N-1)} \trace\left({X}^T (NI - \bm{1}\bm{1}^T ) {X}\right) \\
& = \frac{1}{N(N-1)} \left(E_D^\mathcal{G}({X}) + E_D^{\mathcal{G}^C}({X})\right)
\end{aligned}
\end{equation}

\end{document}